%% file: collas2023_conference.tex
\documentclass{article} 
\usepackage{collas2023_conference,times}

\input{math_commands.tex}

\usepackage{hyperref}
\hypersetup{
	colorlinks=true,
	linkcolor=red,
	filecolor=magenta,
	urlcolor=blue,
	citecolor=purple,
	pdftitle={Overleaf Example},
	pdfpagemode=FullScreen,
}   
\usepackage{graphicx}
\usepackage{subcaption}
\usepackage{amsmath}
\usepackage{amssymb}
\usepackage{booktabs}
\usepackage{url}
\usepackage{multirow}
\usepackage[capitalize]{cleveref}
\crefname{section}{Sec.}{Secs.}
\Crefname{section}{Section}{Sections}
\Crefname{table}{Table}{Tables}
\crefname{table}{Tab.}{Tabs.}

\newcommand{\vgamma}{{\bm \gamma}}
\newcommand{\enquote}[1]{"#1"}
\title{Adiabatic replay for continual learning}

\author{Alexander Krawczyk \\
	Department of Applied Computer Science\\
	University of Applied Sciences Fulda\\
	Leipziger Str. 123, 36037 Fulda, Germany \\
	\texttt{\{alexander.krawczyk\}@cs.hs-fulda.de} \\
	\And 
	Alexander Gepperth  \thanks{\url{www.gepperth.net/alexander}}  \\
	Department of Applied Computer Science\\
	University of Applied Sciences Fulda\\
	Leipziger Str. 123, 36037 Fulda, Germany \\
	\texttt{\{alexander.gepperth\}@cs.hs-fulda.de}
}

\preprintcopy 

\begin{document}
	\maketitle
	\begin{abstract}
		Conventional replay-based approaches to continual learning (CL) require, for each learning phase with new data, the replay of samples representing all of the previously learned knowledge in order to avoid catastrophic forgetting. Since the amount of learned knowledge grows over time in CL problems, generative replay spends an increasing amount of time just re-learning what is already known. In this proof-of-concept study, we propose a replay-based CL strategy that we term adiabatic replay (AR), which derives its efficiency from the (reasonable) assumption that each new learning phase is \textit{adiabatic}, i.e., represents only a small addition to existing knowledge.
		Each new learning phase triggers a sampling process that selectively replays, from the body of existing knowledge, just such samples that are similar to the new data, in contrast to replaying all of it. Complete replay is not required since AR represents the data distribution by GMMs, which are capable of selectively updating their internal representation only where data statistics have changed. As long as additions are adiabatic, the amount of to-be-replayed samples need not to depend on the amount of previously acquired knowledge at all. We verify experimentally that AR is superior to state-of-the-art deep generative replay using VAEs.
	\end{abstract}
	\section{Introduction}\label{sec:intro}
\begin{figure}[h]
		\centering
		\includegraphics*[width=0.4\linewidth,page=4,viewport=0cm 0cm 14.3cm 5cm]{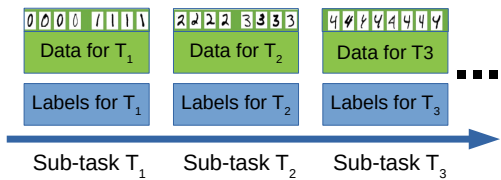}
		\caption{The default scenario for (supervised) CL. The data stream is assumed to be partitioned into \textit{sub-tasks} $T_i$. Statistics within a sub-task are considered stationary, and sub-task data and labels (targets) are assumed to be disjoint, i.e., from different classes (MNIST classes used here for visualization). Sub-task onsets are assumed to be known as well.
			\label{fig:default}
		}
	\end{figure}

\begin{figure}[t]
		\centering
		\includegraphics*[width=0.6\linewidth,page=1,viewport=0in 0in 7.5in 2in]{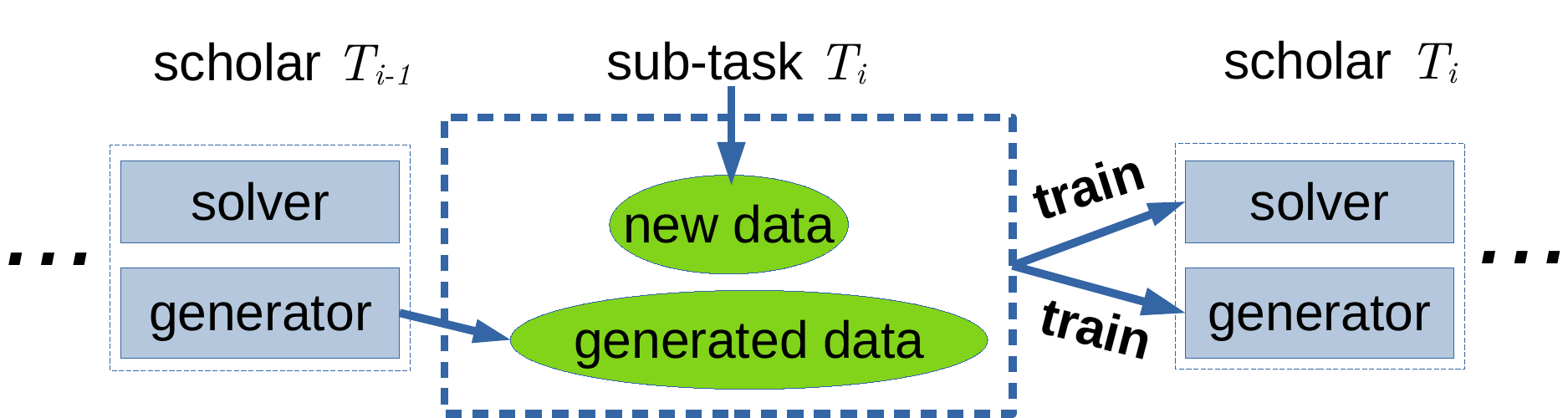}
		\caption{In generative replay, a \textit{scholar} is trained at every sub-task. Scholars are composed of generator and solver modules. The solver performs the task, e.g., classification, whereas the generator serves as a memory for samples from previous sub-tasks $T_{i'}$, $i' < i$. Please note that the amount of generated data usually far exceeds the amount of new data. }
		\label{fig:genrep}
	\end{figure}

	This contribution is in the context of continual learning (CL), a recent flavor of machine learning that investigates learning from data with a non-stationary distribution.
	A common effect in this context is catastrophic forgetting (CF), an effect where previously acquired knowledge is abruptly lost after a change in the data distribution.
	In the default scenario for CL (see, e.g., \cite{bagus2022beyond}), a number of assumptions are made in order to render CL more tractable: first of all, distribution changes are assumed to be abrupt, partitioning the data stream into \textit{sub-tasks} of stationary distribution. Then, sub-task onsets are supposed to be known, instead of inferring them from data. And lastly, sub-tasks are assumed to be disjoint, i.e., not containing the same classes in supervised scenarios. Please refer to \cref{fig:default} for a visualization of the default scenario. Together with this goes the constraint that no, or only a few, samples may be stored.
	A very promising line of work on CL in the default scenario are replay strategies \cite{van2020brain}. \textit{Replay} aims at preventing CF by using samples from previous sub-tasks to augment the current one. On the one hand, there are \enquote{true} replay methods which use a small number of stored samples for augmentation. On the other hand, there are \textit{pseudo-replay} methods, where the samples to augment the current sub-task are produced in unlimited number by a generator, which removes the need to store samples. A schematics of the training process in generative replay is given in \cref{fig:genrep}.
	Replay seems to represent a promising principled approach to CL, but it nevertheless presents several challenges: 
	First of all, if DNNs are employed as solvers and generators, then all classes must be represented in equal proportion at every sub-task.
	For example: in the simple case of a single new class ($D$ samples) per sub-task, the generator must produce $(s-1)D$ samples at sub-task $T_s$ in order to always train with $D$ samples per class. 
	%
	%
	This unbounded and unsaturated linear growth of to-be-replayed samples, and therefore of training time, poses enormous problems for long-term CL.
	By this linear dependency, even very small additions to a large body of existing knowledge (a common use-case) require a large amount of samples to be replayed, see, e.g., \cite{dzemidovich2022empirical}. Since replay is always a lossy process, this imposes a severe limit on GR-based CL performance.
	A related issue is the fact that DNN solvers and generators must be re-trained \textit{from scratch} at every sub-task. This destroys knowledge of previous sub-tasks, which might otherwise be used to speed up training.
	In addition, we are at the mercy of the generator to produce samples in equal frequencies, although this can me mitigated somewhat by class-conditional generation of samples \cite{lesort2019marginal}.
	\subsection{Approach: AR}\label{sec:approach}
	Adiabatic replay (AR), prevents an ever-growing number of replayed samples by re-using existing knowledge, which is only modified where there is a conflict, see \cref{fig:var}. Replay is thus restricted to \textit{conflict regions of data space} by a new selective replay technique.
	To achieve \textit{selective replay} and the re-use of existing knowledge, we rely on fully probabilistic models of machine learning, in particular SGD-trained Gaussian Mixture Models (GMMs) as introduced in \cite{gepperth2021gradient}. We use a single GMM, i.e. a \enquote{flat} architecture with limited modeling capacity, but they are sufficient for the proof-of-concept that we require here. Deep Convolutional GMMs, see \cite{gepperth2021image} are a potential remedy for the capacity problem.
	%
	%
	GMMs are characterized by an explicit similarity (probability) measure which can be used to compare incoming data to existing knowledge as represented by their \textit{components} (i.e., individual centroids, weights and covariance matrices). %
	It was demonstrated in \cite{gepperth2021gradient} that GMMs have intrinsic CL capacity when re-trained with new data. In this case, only the components that are similar to, and thus in conflict with, incoming data are adapted. In contrast, dissimilar components are not adapted, and are thus protected. 
	Each GMM component can perform sampling. In AR, incoming data is augmented by samples drawn from components in conflict regions, which are essentially those with high similarity, see \cref{fig:var}. This ensures that these components are adapted but not simply replaced.
	\subsection{Contributions}
	The salient points of AR are:
	\par\noindent\textbf{Selective replay:} Existing knowledge is not replayed indiscriminately, but only where significant overlap with new data exists. The detection of overlaps is an intrinsic property of the employed GMMs.
	\par\noindent\textbf{Selective replacement:} Existing knowledge is only modified where an overlap with new data exists. Selective replacement is an intrinsic property of GMMs.
	\par\noindent\textbf{Near-Constant time complexity:} Since only a small part of existing knowledge needs to be considered for replay at each sub-task, the number of generated/replayed samples can be small as well. The ratio between new and generated data can therefore be fixed, irrespectively of past sub-tasks.
	\subsection{Related Work}
	In recent years diverging strategies were presented to mitigate CF in CL scenarios.
	\par\noindent\textbf{Regularization}
	These methods add regularization terms to the loss function, aiming to protect knowledge from past tasks, when learning from novel data distributions. They can be split into data-focused approaches, e.g., \cite{li2017learning,jung2016less,rannen2017encoder,maltoni2019continuous,zhang2020class}, where the essential technique is based on knowledge distillation \cite{hinton2015distilling}, and prior-focused approaches, e.g. \cite{kirkpatrick2017overcoming, liu2018rotate, schwarz2018progress, lee2017overcoming, zenke2017continual, aljundi2018memory, aljundi2019task, chaudhry2018riemannian, nguyen2017variational, zeno2018task, ahn2019uncertainty}. Such solutions estimate the distribution over model parameters, used as a prior for learning from new data. The parameter \enquote{importance} is measured, and as a consequence, changes to most relevant parameters are penalized during later training stages. An overview about regularization-based methods can be found in \cite{delange2021continual}. 
	\par\noindent\textbf{Parameter isolation} 
	Dynamically expandable architectures \cite{rusu2016progressive, aljundi2017expert,xu2018reinforced} e.g., inspired by neurogenesis \cite{schwarz2018progress, masse2018alleviating}, rely on extensive memory usage. 
	Parameter isolation with fixed-network capacity, e.g. \cite{fernando2017pathnet, mallya2018packnet, serra2018overcoming, mallya2018piggyback}, depend upon masking during new task training, involving so-called task-oracles for inference, thus restricting this approach to multi-headed setups only.
	\par\noindent\textbf{Rehearsal} 
	This branch of CL solutions relies on the storage of previously encountered data instances. In its purest form, past data is held inside a buffer and mixed with data from the current sub-task to avoid CF, as shown in \cite{gepperth2016bio, rebuffi2017icarl, rolnick2019experience, de2021continual}. This has drawbacks in practice, since it breaks the constraints for task-incremental learning \cite{van2019three}, has privacy concerns, and requires significant memory. \textit{Partial replay}, e.g. \cite{aljundi2019online}, and constraint-based optimization \cite{lopez2017gradient,chaudhry2018efficient,chaudhry2019tiny,aljundi2019gradient}, focus on selecting a subset from previously seen samples, but it appears that the selection of a sufficient subset is still challenging \cite{prabhu2020gdumb}. Comprehensive overviews about current advances in replay can be found in \cite{hayes2021replay,bagus2021investigation}.
	\par\noindent\textbf{Deep Generative Replay} 
	With generative replay (GR), a generator is trained and used for memory consolidation, by replaying samples from previous sub-tasks, see \cref{fig:genrep}. GR using shallow networks was introduced in \cite{robins1995catastrophic}. 
	Deep Generative Replay (DGR) \cite{shin2017continual} employs (deep) generative models like GANs \cite{goodfellow2014generative} and VAEs \cite{kingma2013auto}, combined with DNN solvers.
	The recent growing interest in GR brought up a variety of architectures, either utilizing VAEs \cite{kamra2017deep, lavda2018continual, ramapuram2020lifelong, ye2020learning, caselles2021s}, GANs \cite{atkinson2018pseudo, rios2018closed, wu2018memory, he2018exemplar, ostapenko2019learning, wang2021ordisco, atkinson2021pseudo}, or even such inspired by CLS-theory \cite{kemker2017fearnet, parisi2018lifelong, van2020brain} and many more \cite{titsias2019functional, von2019continual, rostami2019complementary, abati2020conditional, liu2020mnemonics}.
	GAN-based DGR is not considered in this article, since GANs, despite steady improvements, show several shortcomings in the context of CL, see \cite{dzemidovich2022empirical}: First of all, GANs are subject to mode collapse, which is hard to detect in an automated way, since they do not optimize an easy-to-compute loss function. Furthermore, GANs learn to generate data \textit{from} Gaussian noise, but they do not map the data \textit{to} Gaussian noise, skipping an explicit density estimation, and thus are incapable of performing inference queries, e.g., outlier detection. Lastly, as stated in \cite{richardson2018gans}, GANs do not seem to model the intrinsic distribution of the data without important extensions, e.g., Wasserstein distance \cite{arjovsky2017wasserstein}, which increases computational burden and may be biased when it comes to replaying data for CL.
	\par\noindent\textbf{VAE-based DGR} 
	Without prior data access to fine-tune GAN-based generators, VAEs seem to be a better choice as a generator, since they avoid mode collapse under resource constraints \cite{dzemidovich2022empirical}. VAEs model the probability distribution of the data, albeit in an indirect fashion, by the constrained optimization of a reconstruction loss (variational lower bound). 
	\par\noindent\textbf{GMR} Gaussian Mixture Replay (GMR), proposed in \cite{pfulb2021overcoming}, is a generative replay approach and serves as the foundation for AR.
	GMR combines GMMs as generators, enabling an explicit density estimation, combined with a solver into a single-fused model. 
	Our approach is similar to GMR in terms of the architectural design. However, we go beyond GMR in the organization of replay, which aims for constant time-complexity w.r.t number of processed training tasks.
	%
	\section{Methods}\label{sec:methods}
	\subsection{AR: Fundamentals}\label{sec:gmm}
	\begin{figure}[t!]
		\centering
		\includegraphics*[width=0.6\linewidth,page=2,viewport=0in 0in 5.6in 2.5in]{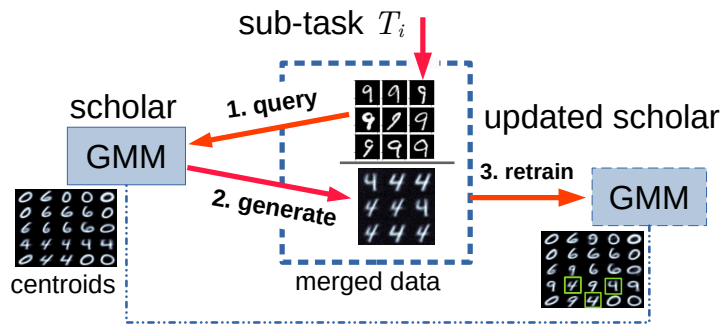}
		\caption{The proposed AR approach, illustrated in an experimental setting. The scholar (GMM) has been trained on MNIST classes 0, 4 and 6 in sub-task $T_1$. At sub-task $T_2$, new data (class 9) is used to \textit{query} the scholar for similar samples, resulting in the selective replay of mostly 4's but no 0's. The scholar is re-trained \textit{from its current state}, so no data concerning class 0 is required. Re-training results in the insertion of 9's into the existing components.}
		\label{fig:var}
	\end{figure}
	\par\noindent\textbf{Big picture} In contrast to conventional replay, where a scholar is composed of a generator and a solver network, see \cref{fig:genrep}, AR proposes scholars with a single network performing both functions.
	Assuming a suitable scholar (see below), the high-level logic of AR is shown in \cref{fig:var}:
	Each sample from a new sub-task is used to \textit{query} the scholar, which generates a similar, known sample. Mixing new and generated samples in a defined, constant proportion creates the training data for the current sub-task.
	A new sample will cause adaptation of the scholar in a localized region of data space. Variants generated by that sample will, due to similarity, cause adaptation in the same region. Knowledge in the overlap region will therefore be adapted to represent both, while dissimilar regions stay unaffected (see \cref{fig:var} for a visual impression).
	\par\noindent\textbf{Requirements on the scholar} are twofold: first of all, it must be able to query its internal representation to selectively generate similar samples. Furthermore, it must be able to selectively adapt the parts of its internal representation  that coincide with new and generated samples, and to leave the remaining part untouched. None of these requirements are fulfilled by DNNs, which is why we implement the scholar by a \enquote{flat} GMM layer followed by a linear classifier. All layers are independently trained via SGD according to \cite{gepperth2021gradient}. Extensions to deep convolutional GMMs (DCGMMs) \cite{gepperth2021new} for higher sampling capacity can be incorporated as drop-in replacements.
	\par\noindent\textbf{GMMs} describe data distributions by a set of $K$ \textit{components}, consisting of component weights $\pi_k$, centroids $\vmu_k$ and covariance matrices $\mSigma_k$. A data sample $\vx$ is assigned a probability $p(\vx) = \sum_k \pi_k \mathcal N(\vx ; \vmu_k, \mSigma_k)$ as a weighted sum of normal distributions $\mathcal N(\vx; \vmu_k, \mSigma_k)$. Training of GMMs is performed as detailed in \cite{gepperth2021gradient} by adapting centroids, covariance matrices and component weights through the SGD-based minimization of the negative log-likelihood:
	\begin{equation}
		\mathcal L =\sum_n \log \sum_k \pi_k \mathcal N(\vx_n; \vmu_k,\mSigma_k).	
	\end{equation}
	\par\noindent\textbf{Variant generation} is a form of sampling from the probability density represented by a trained GMM, see \cite{gepperth2021image}.
	It is triggered by a query in the form of a data sample $\vx_n$, which is converted into a control signal $\mathcal{T}$ defined by the posterior probabilities (or \textit{responsibilities}):
	\begin{equation}
		\gamma_k(\vx_n) = \frac{\pi_k\mathcal N(\vx_n; \vmu_k,\mSigma_k)}{\sum_j \pi_j \mathcal N(\vx_n; \vmu_j,\mSigma_j)}.
	\end{equation}
	The responsibilities parameterize a multinomial distribution for drawing a GMM component $k^*$ to sample from, which is simple because it defines a multi-variate normal distribution $\hat \vx \sim \mathcal N(\cdot; \vmu_{k^*}, \mSigma_{k^*})$.
	To reduce noise, top-S sampling is introduced, where only the $S$ highest values in the control signal are used for selection.
	\par\noindent\textbf{Classification} is performed by feeding GMM responsibilities into a bias-free, linear regression layer as $\vo(\vx_n) = \mW\vgamma(\vx_n)$.
	We use a MSE loss and drop the bias term to reduce the sensitivity to unbalanced classes, which is characteristic of adiabatic replay.
	\par\noindent\textbf{GMM training} uses the procedure described in \cite{gepperth2021gradient}. An important quantity in this procedure is the annealing radius $r(t)$, which is automatically reduced from its start value $r_0$ over time.
	\subsection{AR: Implementation}\label{sec:implementation}
	AR employs a GMM layer $L_{(G)}$ with $K=400$ components and diagonal covariance matrices as its scholar. The choice of $K$ is subject to a \enquote{the more the better} principle, and is limited only by available GPU memory. We aim to use the extension to DCGMMs in the future to replace the flat $L_{(G)}$ by a sophisticated generative structure.
	We follow the best-practice strategies presented and justified in \cite{gepperth2021gradient}. We keep the same model hyper-parameters throughout all experiments, since their choice is independent of the CL problem at hand. One exception is adjusting $\gamma=0.96$ (GMM annleaing control) for the initial (from scratch) training on $T_1$ to support component convergence, while reducing it back to $\gamma=0.9$ for the subsequent replay-tasks $T_{i>1}$.
	Both the GMM and the linear classifier are independently trained via vanilla SGD with a fixed learning rate of $\epsilon=0.05$.
	Annealing controls the component adaptation radius for $L_{(G)}$ via parameter $r_{0}$. It is set to $r_{0}^{init}=\sqrt{0.125K}$ for the first task $T_1$, and $r_{0}^{replay}=0.1$ for $T_{i>1}$. $r_{0}$ is reset to these fixed values before each training.
	GMM sampling parameters $S=3$ (top-S) and $\rho=1.0$ (normalization) are kept fixed throughout all experiments.
	We provide a publicly available TensorFlow 2 \cite{abadi2016tensorflow} implementation\footnote{The repository can be accessed \href{https://github.com/Alexk1704/sccl}{\textcolor{blue}{here}}. The code will be made publicly available for the camera-ready version. Detailed instructions will guide the user through experiment reconstruction. The executables come with default parameters as presented in this section.}
	\subsection{Deep Generative Replay: Implementation}
	\begin{table}[h]
		\setlength{\arrayrulewidth}{0.25mm}
		\renewcommand{\arraystretch}{1.25}
		\centering
		\begin{tabular}{l}
			\multicolumn{1}{c}{\textbf{VAE Components}}\\
			\hline
			\textbf{Encoder}: \\
			{Conv2D(32,5,2)-ReLU-Conv2D(64,5,2)-} \\
			{ReLU-Dense(100)-ReLU-Dense(50)-ReLU-Dense(25)} \\
			\hline	
			\textbf{Decoder}: \\
			{Dense(100)-ReLU-Dense(3136)-} \\
			{Reshape(16,14,14)-ReLU-Conv2DTranspose(32,5,2)-} \\
			{ReLU-Conv2DTranspose(1,5,2)-Sigmoid} \\
			\hline
			\textbf{Solver}: \\
			{Conv2D(32,5,1)-MaxPool2D(2)-ReLU-} \\
			{Conv2D(64,5,1)-MaxPool2D(2)-ReLU-Dense(100)-} ReLU-Dense(10) \\
		\end{tabular}
		\caption{DNN structure for VAE-based replay.
			\label{tab:networkstructure}
		}
	\end{table}
	We implement deep generative replay (DGR) using VAEs as generators, see \cite{dzemidovich2022empirical}. The VAE latent dimension is 25, the disentangling factor $\beta=1.$, and conditional sampling is turned off.
	The structure of the generator and solver networks is given in \cref{tab:networkstructure}. We choose VAEs over GANs or WGANs due to the experiments conducted in \cite{dzemidovich2022empirical}, which suggest that GANs require extensive structural tuning, which is by definition excluded in a CL scenario for all sub-tasks but the first. Similarly, GANs and VAEs were both used in CL research, e.g., in \cite{lesort2019marginal} with comparable performance. 
	The learning rate for solvers and VAE generators are $\epsilon_S=10^{-4}$, $\epsilon_{VAE}=10^{-3}$ using the Adam optimizer.
	\subsection{Performance Criteria}\label{sec:exppeval}
	This work largely adopts the notation of \cite{mundt2021cleva}. We provide common accuracy-based metrics such as, e.g., presented in \cite{kemker2018measuring}, to measure the classification performance of a scholar $\mathcal{M}_{t}$ after being trained on each sub-task $t$ of a CIL-P from \cref{tab:slts}. Moreover, we present common forgetting measures \cite{chaudhry2018efficient} for F-MNIST and E-MNIST.
	Per-task classification accuracy is provided in $T\times T$ matrices $\alpha \in \mathbb{R}^{T \times T}$, whereas each entry $a_{i,j}$ denotes the test set accuracy of $\mathcal{M}_{j}$ after processing sub-task $j$.
	%
	%
	Additionally, the offline performance $\alpha^{base}_{T}$ is measured on the baseline test set $D_{all}$, consisting of test samples from each sub-task of the respective CIL-P after $T$ sub-tasks were seen.
	We denote the initial accuracy on the first sub-task $T_1$ after training on $T_1$ by $\alpha^{init}$, while $\alpha^{init}_{T}$ reflects performance on $T_1$ after $T$ sub-tasks. All listed accuracy values are normalized to a range of $\alpha \in [0,1]$.
	%
	%
	%
	%
	Forgetting $F_{i}^{j}$ is defined for a sub-task $i$ after $\mathcal{M}$ was trained on (including) $j$. It reflects the loss of knowledge about a previous sub-task $i$ and measures the difference to peak performance of $\mathcal{M}$ on exactly that task:
	\begin{equation}
		F_{i}^{j} = \max_{i\in\{1,..,t-1\}} \alpha_{i,j} - \alpha_{t,j} \qquad \forall j < t.
	\end{equation}
	Average forgetting $F_t$ is thus defined as:
	\begin{equation}
		F_t = \frac{1}{T-1} \sum^{T-1}_{j=1} F^{t}_{j}.
	\end{equation}
	Backwards-Transfer $BWT_{t,j}$ \cite{lopez2017gradient, chaudhry2018riemannian}, is highlighted as negative forgetting $-F_{i}^{j}$. The measures are organized into a $T-1 \times T$ evaluation matrix, as e.g., shown in \cite{bagus2021investigation}. Additionally, we provide data to capture the task-similarity for future sub-tasks $j>t$, which shows AR's intrinsic capacity to detect \enquote{data overlap}, by providing GMM log-likelihoods after learning a sub-task $t$, see \cref{fig:gen_samples_loglik_plot}.
	\section{Experiments}\label{sec:exp}
	\subsection{Evaluation Data}\label{sec:data}
	\noindent\textbf{MNIST}~\cite{lecun1998gradient} consists of $60\,000$ $28$\,$\times$\,$28$ gray scale images of handwritten digits (0-9).\\[3pt]
	\noindent\textbf{Fashion-MNIST}~\cite{xiao2017fashion} consists of images of clothes in 10 categories and is structured like MNIST. \\[3pt]
	\noindent\textbf{E-MNIST}~\cite{cohen2017emnist} is structured like MNIST and extends it by letters. We use the balanced split which contains 131.000 samples in 47 classes.
	\\[3pt]
	These datasets are not particularly challenging w.r.t. non-continual classification, although CL tasks formed from these datasets according to the default scenario (\cref{sec:intro}) have been shown to be very difficult, see, e.g., \cite{pfulb2019comprehensive}. Particular examples of CL tasks formed from MNIST and Fashion-MNIST that have been used in the literature are D9-1, D5$^2$, D2$^5$ or D1$^{10}$. Expressions like D$2^5$ are to be read as D2-2-2-2-2.
	E-MNIST enables a CL scenario where the amount of already acquired knowledge can be significantly larger than the amount of new data added with each successive sub-task. Therefore, D20-$1^5$ is performed exclusively for E-MNIST.
	\subsection{Experimental Setting}
	\par\noindent\textbf{Experiments}
	Were run on a machine with 32 GB RAM, and a single RTX3070Ti GPU.
	Replay is investigated in a supervised CIL-scenario, assuming known task-boundaries and disjoint classes.
	Sub-tasks $T_{i}$ contain all samples of the corresponding classes defining them, see \cref{tab:slts} for details. It is assumed that data from all tasks occurs with an equal probability.
	Training consists of an (initial) run on $T_1$, followed by a sequence of independent (replay) runs on $T_{i>1}$.
	We perform ten randomly initialized runs for each CIL-Problem, and conduct baseline experiments for all datasets, measuring the offline performance.
	\begin{table}[h!]
		\small
		\renewcommand{\arraystretch}{1.25}
		\centering
		\begin{tabular}{ c | c | c | c | c | c | c }
			\large{CIL-Problem $\downarrow$} & \large{$T_1$} & \large{$T_2$} & \large{$T_3$} & \large{$T_4$} & \large{$T_5$} & \large{$T_6$} \\[0.5ex]
			\hline
			D5-$1^5$A & [0-4] & 5 & 6 & 7 & 8 & 9 \\
			D5-$1^5$B & [5-9] & 0 & 1 & 2 & 3 & 4 \\
			\hline
			D7-$1^3$A & [0-6] & 7 & 8 & 9 & / & / \\
			D7-$1^3$B & [3-9] & 0 & 1 & 2 & / & / \\
			\hline
			D20-$1^5$A & [0-19]	& 20 & 21 & 22 & 23 & 24 \\
			D20-$1^5$B & [5-24] & 0 & 1 & 2 & 3 & 4 
		\end{tabular}
		\caption{Showcase of investigated CL/CIL-Problems. Each sub-task $T_i$ contains the complete image and label data $(X,Y)$ from the corresponding classes. Initial task $T_1$ data is balanced w.r.t classes. D20-$1^5$A and D20-$1^5$B are exclusive for E-MNIST.
			\label{tab:slts}
		}
	\end{table}
	\par\noindent\textbf{Training} We set the mini-batch size to $\beta=100$, and define a mixing strategy for replay-training. $\beta_G$ represents the amount of generated/artificial samples from a current scholar $S_{i-1}$, while $\beta_R$ denotes the amount of \enquote{real} data from an incoming sub-task $T_{i}$, such that $\beta = \beta_G + \beta_R$. Therefor, mini-batches are randomly drawn from a merged subset, consisting of generated and real data.
	The mixing strategy is defined by the parameter $\chi = \frac {\beta_G} {\beta_R}$ which controls replay. We define $D_i$ as the number of samples in $T_i$. While, $\chi(D_i)$ denotes the generated data, to be mixed with $D_i$, to form a merged training set $\mathcal{D}_{T_i}$. Two important strategies depending on $\chi$ are explained in great detail:
	\\
	\par\noindent\textbf{Balanced} This strategy ensures a variable-length training set for each sub-task $T_i$, which grows linearly with $i$. This increases the number of generated samples without any restrictions $\sum_{j=1}^{i-1} D_j$, see \cref{sec:intro}, and facilitates a class-balancing mechanism w.r.t each encountered sub-task. According to this constraint, it becomes mandatory to track information about class occurrence. E.g., a scholar $S_{1}$ learning a \enquote{9-1 task}, setting $\chi=9$, results in $\beta_G=90$ and $\beta_R=10$. This assumes that the generator naturally produces past data in equal proportions. This can be enforced by conditional sampling, but is ultimately left to the model itself.
	\\
	\par\noindent\textbf{Constant-time} Here, we limit the number of generated samples to $D_i$ for replay, where $D_i$ is the amount of samples contained in each sub-task. Setting $\chi=1$ for all sub-tasks is a trivial approach to dismiss all assumptions regarding task/class balancing for the resulting merged data. Please note that we do not assume an equal distribution of past sub-task classes, nor is it required to preserve any information about previously encountered data instances/labels. This strategy keeps the number of generated samples constant with each sub-task, and thus comes with modest temporary storage requirements.
	\\ \\
	VAE-solvers are trained for 20 epochs, where as VAE-generators are trained for 50 epochs on merged task-data $\mathcal{D}_{T_{i}}$. We expose DGR to both training scenarios, while AR experiments are conducted for the constant-time setting only.
	Training iterations, i.e., the amount of steps over the constituted mini-batches $\beta$, are calculated dynamically for each sub-task. This affects the balanced mixing strategy, as $D_i$ grows linearly, affecting the training duration negatively. 
	AR follows the procedure presented in \cite{gepperth2021gradient}, and implements early stopping, terminating training when GMM $L_{(G})$ reaches a plateau of stationary loss for $T_i$. We limit training to 128 epochs for initial runs, and resp. 256 epochs for replay-training. 
	\subsection{Variant generation with GMMs}
	\begin{figure}[h!]
		\centering
		\includegraphics*[width=0.25\linewidth,viewport=10pt 0pt 450pt 350pt,draft=False]{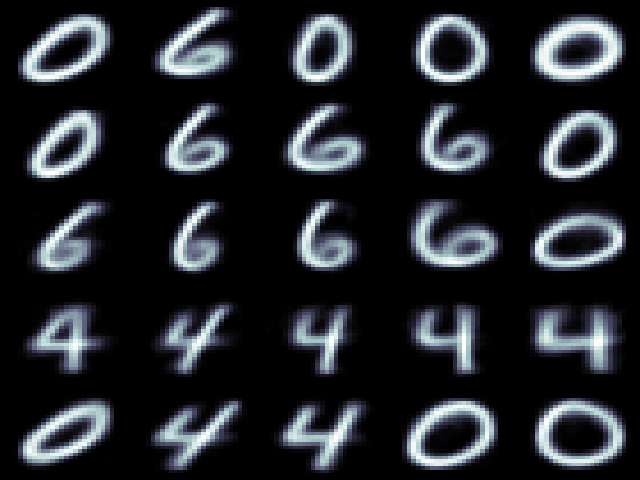}
		\hspace{0.1cm}
		\includegraphics*[width=0.20\linewidth,viewport=10pt 10pt 180pt 190pt,draft=False]{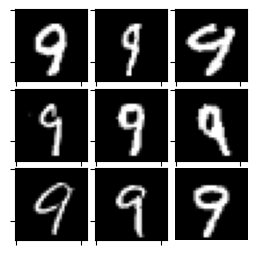}
		\hspace{0.1cm}
		\includegraphics*[width=0.234\linewidth,viewport=30pt 0pt 430pt 340pt,draft=False]{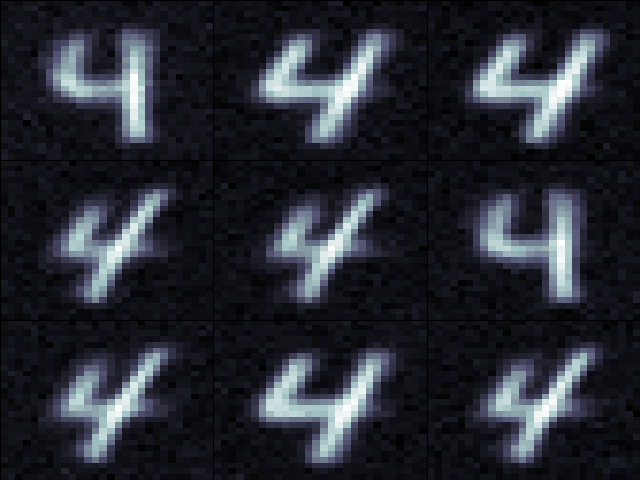}
		\caption{\label{fig:vargen} An example for variant generation in AR, see text and \cref{fig:var} for details. Left: centroids of the current GMM scholar trained on MNIST classes 0,4,6. Middle: query samples of MNIST class 9. Right: variants generated in response to the query. Component weights and variances are not shown.
		}
	\end{figure}
	We illustrate the capacity of GMM layer $L_{(G)}$ to query its internal representation by data samples, and to selectively generate samples that are \enquote{best matching} to those defining the query. To show this, we train a GMM layer of $K=25$ components on MNIST classes 0,4 and 6 for 50 epochs using the best-practice rules from \cite{gepperth2021gradient}. Then, we query the trained GMM with samples from class 9 uniquely as described in \cref{sec:gmm}. The resulting samples are all from class 4, since it is the class that is most similar to the query class. These results are visualized in \cref{fig:var}. Variant generation results for deep convolutional extensions of GMMs can be found in \cite{gepperth2021new}, showing that the AR approach can be scaled to more complex problems.
	\subsection{Comparison: AR and DGR-VAE}
	From the collected experimental data, we deduce that CIL is still challenging, especially when the replay mechanism is restricted to enforce a constant time- and space-complexity regarding a growing amount of sequential sub-tasks with small data additions. 
	\\ \\
	DGR shows its superiority across all datasets and CIL-problems from \cref{tab:short_results} by achieving higher performance on the baseline $\alpha^{base}$, as well as the initial-task $\alpha^{init}$. However, this is not a surprising fact considering the comparison of model complexity, as DGR-VAE has 2.15 times more trainable parameters than AR in it's current state. DGR-VAE profits from this increased representational power, and also the fact that no replay is required for the joint-class training on $T_1$.
	Nevertheless, DGR-VAE shows strongly divergent results for replay-training $T_{>=2}$, depending on the investigated training scenario (balanced/constant). We will discuss these differences in great detail:
	\begin{table}[h]
		\setlength{\arrayrulewidth}{0.15mm}
		\renewcommand{\arraystretch}{1.5}
		\setlength{\tabcolsep}{4pt}
		\centering
		\tiny
		\begin{tabular}{c l | c | c | c | c | c | c}
			\multirow{2}{*}{\large{CIL-P.}} & \multirow{2}{*}{\Large{$\alpha_i^j$}} 
			& \multicolumn{3}{|c|}{\large{\textbf{MNIST}}} & \multicolumn{3}{|c}{\large{\textbf{Fashion-MNIST}}} \\[3pt]
			& & \large{\textbf{AR}} & \large{DGR(c.)} & \large{DGR(b.)} 
			& \large{\textbf{AR}} & \large{DGR(c.)} & \large{DGR(b.)} \\[3pt]
			\hline
			\large{$D_{all}=[0,10]$} & $\alpha^{base}$  
			& $.92$  & \multicolumn{2}{c|}{\textbf{.98}}
			& $.78$  & \multicolumn{2}{c}{\textbf{.89}} \\[1.5pt]
			\hline                   						
			\multirow{4}{6.5em}{\large{D5-$1^5$A}}  
			& $\alpha^{init}$  				
			& $.96$	& \multicolumn{1}{c|}{\textbf{.99}} & \multicolumn{1}{c|}{\textbf{.99}}	
			& $.82$	& \multicolumn{1}{c|}{\textbf{.91}} & \multicolumn{1}{c}{\textbf{.91}} \\[1.5pt]
			& $\alpha^{init}_{T6}$ 		
			& $.85$	& \multicolumn{1}{c|}{$.40$} & \multicolumn{1}{c|}{\textbf{.78}}
			& $.75$	& \multicolumn{1}{c|}{$.53$} & \multicolumn{1}{c}{\textbf{.82}} \\[1.5pt]
			& $\alpha^{base}_{T6}$  		
			& $.76$	& \multicolumn{1}{c|}{$.63$} & \multicolumn{1}{c|}{\textbf{.83}}
			& $.70$	& \multicolumn{1}{c|}{$.63$} & \multicolumn{1}{c}{\textbf{.77}} \\[1.5pt]
			& $F_{T6}$  					
			& \textbf{.06}  & \multicolumn{1}{c|}{.21} & \multicolumn{1}{c|}{.11}
			& \textbf{.14}	& \multicolumn{1}{c|}{$.17$} & \multicolumn{1}{c}{$.28$} \\[1.5pt]
			\hline
			\multirow{4}{6.5em}{\large{D5-$1^5$B}}  
			& $\alpha^{init}$  				
			& $.95$	& \multicolumn{1}{c|}{\textbf{.98}} & \multicolumn{1}{c|}{\textbf{.98}}
			& $.91$	& \multicolumn{1}{c|}{\textbf{.97}} & \multicolumn{1}{c}{\textbf{.97}} \\[1.5pt]
			& $\alpha^{init}_{T6}$ 		
			& $.82$	& \multicolumn{1}{c|}{$.43$} & \multicolumn{1}{c|}{\textbf{.84}}
			& \textbf{.74}	& \multicolumn{1}{c|}{$.43$} & \multicolumn{1}{c}{\textbf{.74}} \\[1.5pt]
			& $\alpha^{base}_{T6}$  		
			& $.80$	& \multicolumn{1}{c|}{$.70$} & \multicolumn{1}{c|}{$.90$}
			& $.71$	& \multicolumn{1}{c|}{$.56$} & \multicolumn{1}{c}{$.72$} \\[1.5pt]
			& $F_{T6}$  					
			& \textbf{.02}	& \multicolumn{1}{c|}{.12} & \multicolumn{1}{c|}{.05}
			& \textbf{.14}	& \multicolumn{1}{c|}{$.35$} & \multicolumn{1}{c}{$.28$} \\[1.5pt]
			\hline
			\multirow{4}{6.5em}{\large{D7-$1^3$A}}
			& $\alpha^{init}$  				
			& $.97$	& \multicolumn{1}{c|}{\textbf{.99}} & \multicolumn{1}{c|}{\textbf{.99}}
			& $.76$	& \multicolumn{1}{c|}{$.86$} & \multicolumn{1}{c}{\textbf{.87}} \\[1.5pt]
			& $\alpha^{init}_{T4}$ 		
			& $.86$	& \multicolumn{1}{c|}{$.67$} & \multicolumn{1}{c|}{\textbf{.91}}
			& $.70$	& \multicolumn{1}{c|}{$.60$} & \multicolumn{1}{c}{\textbf{.77}} \\[1.5pt]
			& $\alpha^{base}_{T4}$  		
			& $.81$	& \multicolumn{1}{c|}{$.75$} & \multicolumn{1}{c|}{\textbf{.90}}
			& $.75$	& \multicolumn{1}{c|}{$.69$} & \multicolumn{1}{c}{\textbf{.80}} \\[1.5pt]
			& $F_{T4}$  					
			& \textbf{.05}	& \multicolumn{1}{c|}{.12} & \multicolumn{1}{c|}{.08}
			& \textbf{.05}	& \multicolumn{1}{c|}{$.12$} & \multicolumn{1}{c}{$.08$} \\[1.5pt]
			\hline
			\multirow{4}{6.5em}{\large{D7-$1^3$B}}  
			& $\alpha^{init}$  				
			& $.92$	& \multicolumn{1}{c|}{\textbf{.98}} & \multicolumn{1}{c|}{\textbf{.98}}
			& $.85$	& \multicolumn{1}{c|}{\textbf{.93}} & \multicolumn{1}{c}{\textbf{.93}} \\[1.5pt]
			& $\alpha^{init}_{T4}$ 		
			& $.89$	& \multicolumn{1}{c|}{$.75$} & \multicolumn{1}{c|}{\textbf{.94}}
			& $.71$	& \multicolumn{1}{c|}{$.56$} & \multicolumn{1}{c}{\textbf{.73}} \\[1.5pt]
			& $\alpha^{base}_{T4}$  		
			& $.87$	& \multicolumn{1}{c|}{$.83$} & \multicolumn{1}{c|}{\textbf{.95}}
			& $.73$	& \multicolumn{1}{c|}{$.66$} & \multicolumn{1}{c}{\textbf{.75}} \\[1.5pt]
			& $F_{T4}$  					
			& \textbf{.00} 	& \multicolumn{1}{c|}{.06} & \multicolumn{1}{c|}{.02}
			& \multicolumn{1}{c|}{$.04$}	& \multicolumn{1}{c|}{$.06$} & \textbf{.02} \\[1.5pt]
			\\
			\multirow{2}{*}{\large{...}} & \multirow{2}{*}{\large{...}} & \multicolumn{6}{c}{\large{\textbf{E-MNIST}}} \\[3pt]
			& & \multicolumn{2}{c}{\large{\textbf{AR}}} & \multicolumn{2}{c}{\large{DGR(c.)}} & \multicolumn{2}{c}{\large{DGR(b.)}} \\[3pt]
			\hline
			\large{\textbf{$D_{all}=[0,24]$}} & \large{$\alpha^{base}$}
			& \multicolumn{2}{c}{$.67$} & \multicolumn{4}{|c}{\textbf{.73}} \\[1.5pt]
			\hline                   						
			\multirow{4}{*}{\large{D20-$1^5$A}}
			& $\alpha^{init}$  				
			& \multicolumn{2}{c|}{$.72$} & \multicolumn{2}{|c|}{\textbf{.92}} & \multicolumn{2}{c}{\textbf{.92}} \\[1.5pt]
			& $\alpha^{init}_{T6}$ 		
			& \multicolumn{2}{c|}{$.63$} & \multicolumn{2}{|c|}{.11} & \multicolumn{2}{c}{\textbf{.76}} \\[1.5pt]
			& $\alpha^{base}_{T6}$  		
			& \multicolumn{2}{c|}{$.61$} & \multicolumn{2}{|c|}{.25} & \multicolumn{2}{c}{\textbf{.74}} \\[1.5pt]
			& $F_{T6}$  					
			& \multicolumn{2}{c|}{\textbf{.03}} & \multicolumn{2}{c|}{$.29$} & \multicolumn{2}{c}{$.26$} \\[1.5pt]
			\hline
			\multirow{4}{*}{\large{D20-$1^5$B}}
			& $\alpha^{init}$  				
			& \multicolumn{2}{c|}{$.69$} & \multicolumn{2}{|c|}{\textbf{.92}} & \multicolumn{2}{c}{\textbf{.92}} \\[1.5pt]
			& $\alpha^{init}_{T6}$ 		
			& \multicolumn{2}{c|}{$.60$} & \multicolumn{2}{|c|}{.08} & \multicolumn{2}{c}{\textbf{.74}} \\[1.5pt]
			& $\alpha^{base}_{T6}$  		
			& \multicolumn{2}{c|}{$.59$} & \multicolumn{2}{|c|}{.24} & \multicolumn{2}{c}{\textbf{.72}} \\[1.5pt]
			& $F_{T6}$  					
			& \multicolumn{2}{c|}{\textbf{.04}} & \multicolumn{2}{c|}{$.24$} & \multicolumn{2}{c}{$.29$} \\[1.5pt]
			
		\end{tabular}
		\caption{Experimental results for AR and DGR (in the \textbf{c}onstant-time and \textbf{b}alanced training setting). Metrics are averaged across $N=10$ runs. Detailed information about the evaluation and experimental setup can be found in \cref{sec:exppeval}.
			\label{tab:short_results}
		}
	\end{table}
	\par
\begin{figure}[h!]
		\centering
		\begin{subfigure}{.5\textwidth}
			\centering
			\includegraphics[width=1.\linewidth]{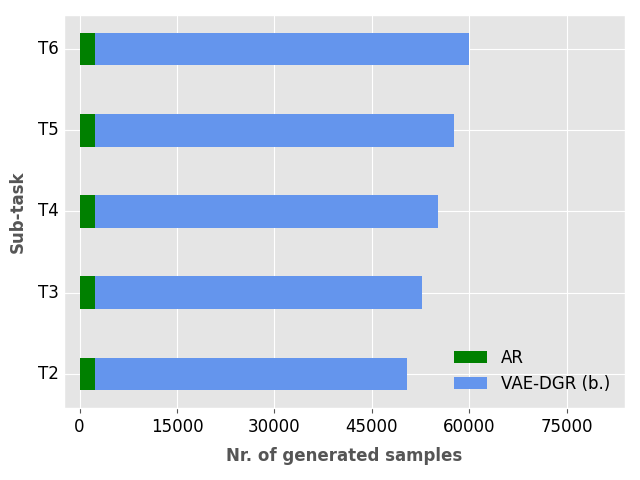}
		\end{subfigure}%
		\begin{subfigure}{.5\textwidth}
			\centering
			\includegraphics[width=1.\linewidth]{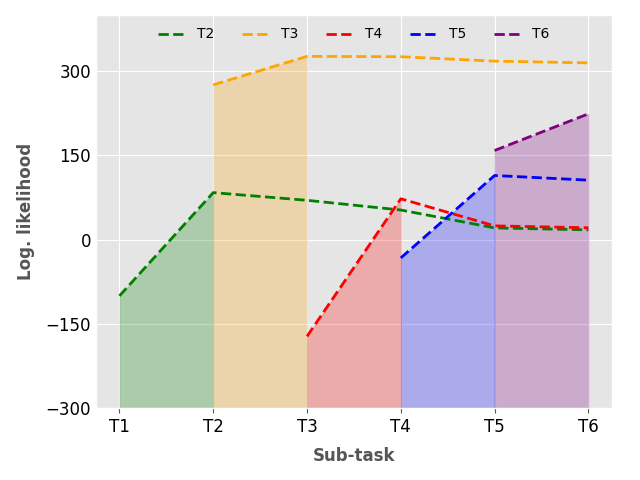}
		\end{subfigure}
		\caption{Example: E-MNIST20-1. Left: Number of generated samples per sub-task for AR/VAE-DGR(b.). Right: Successive sub-task have no significant overlap. This is shown using the negative GMM log-likelihood for AR after training on sub-task $T_i$ and then keeping the GMM fixed. As we can observe, log-likelihood universally drops, indicating a poor match. }
		\label{fig:gen_samples_loglik_plot}
	\end{figure}

\noindent\textbf{Balanced setting}
	Even when a task/class-balancing is ensured, DGR-VAE (b.) suffers from forgetting as can be seen in \cref{tab:app_forgetting_mats}. It appears that the shift in data distribution towards the most recent sub-tasks, and its resulting implications, cannot solely be compensated by an increased amount of generated data.
	While experiments on MNIST do not show any significant forgetting. A sharp drop in performance is observed for the replay-training on $T_2$-$T_4$, see e.g., results on Fashion-MNIST D5-$1^5$A for task $T_2$, D5-$1^5$B for $T_4$, E-MNIST D20-$1^5$A/B for $T_2,T_3$ from \ref{tab:app_dgr-b_accmatrix}. Fashion-MNIST D7-$1^3$A/B shows comparable results, although not as bad, since it resembles a rather short learning problem (with only 3 successive training phases), which may not be enough to encounter strong forgetting.
	This effect appears to be primarily caused by an increased complexity, arising from a high overlap between the represented and new data statistics. We encountered severe knowledge loss for VAE-DGR in both settings, whenever it had to deal with the successive occurrence of individual classes, sharing a high similarity in the input space.
	\\ \\ 
	Noticeably, AR shows some promising results in terms of knowledge preservation, as well as strong forgetting prevention for the learned classes from tasks $T_1$-$T_5$.
	This is best demonstrated by comparing $\alpha^{init}$ (5, 7, or 20 classes) to $\alpha^{init}_{T}$ after performing replay-training on all tasks $T$. We see a comparatively small difference between these values, while identifying this 
	particular metric to be crucial, as it represents the architecture's ability to deal with small incremental additions/updates to the internal knowledge base over a sequence of tasks.
	\\ \\
	The boundlessly expanding storage demand for generated samples $D_i$ with each sub-tasks is a major drawback in terms of scalability towards growing training sequences for DGR-VAE, as shown for the example of E-MNIST D20-$1^5$ in \cref{fig:gen_samples_loglik_plot}.
	Additionally, one should consider the temporal aspect of this setting, since DGR-VAE requires about two to three times as much training time, for e.g., Fashion-MNIST D5-$1^5$, as in a constant setting and/or using AR.
	\par\noindent\textbf{Constant-time scenario}
	When constrained to constant-time, DGR-VAE (c.) is heavily outperformed by AR and DGR-VAE (b.), as it suffers from tremendous accuracy drops, due to severe forgetting across all CIL-P, as presented in \cref{tab:app_dgr-c_accmatrix}. AR demonstrates its intrinsic ability to limit unnecessary overwrites of past knowledge (see \cref{tab:app_ar_accmatrix,tab:app_forgetting_mats}) by performing efficient \textit{adiabatic updates}, as showcased in \cref{fig:vargen}.
	AR exhibits another important property, namely it's modest (temporary) memory requirements, as the amount of generated samples is kept constant w.r.t an ever-increasing number of sub-tasks, see \cref{fig:gen_samples_loglik_plot} for an example.
	AR training is mainly characterized by matching GMM components with arriving input. Therefor performance on previous sub-tasks slightly decreases through the adaptation of selected/similar units. This implies that a GMM tends to converge towards a \textit{trade-off} between past knowledge and new data. This effect is clearly seen in the successive (replay-)training of two \enquote{similar} classes, e.g. seen for, Fashion-MNIST D5-$1^5$A task $T_2$ \enquote{class: sandals} and task $T_4$ \enquote{class: sneakers}.
	\section{Discussion}
	In summary, we can state that our AR approach clearly surpasses VAE-based DGR in the evaluated CIL-P when constraining replay to a constant-time strategy. This is remarkable because the AR scholar performs the tasks of both solver and generator, while at the same time having less parameters. The advantage of AR becomes even more pronounced when considering forgetting prevention instead of simply looking at the classification accuracy results.
	Based on this proof-of-concept, we may therefore conclude that AR offers a principled approach to truly long-term CL. In the following text, we will discuss salient points concerning our evaluation methodology and the conclusions we draw from the results:
	\\
	\par\noindent\textbf{Data} 
	The used datasets are not considered meaningful benchmarks in non-continual ML due to their simplicity. Although, in CL, even MNIST and Fashion-MNIST are still considered as difficult, in particular for VAE-DGR (see \cite{dzemidovich2022empirical,pfulb2019comprehensive}), and many CL studies rely on these two datasets. E-MNIST represents a benchmark that is simple to solve for non-continual classifiers, but quite hard for CL, due to the large number of classes and the possibility to form rather lengthy training periods. This may result in a multitude of successive tasks, only adding a small fraction of knowledge with each learning period. Considering the experimental results, we encourage to move towards more practically realistic CL-problem formulation, to truly evaluate CF resistance in such scenarios.
	\par\noindent\textbf{Fairness of comparison}
	Considering the number of free parameters, the comparison is biased in favor of VAE-DGR, which makes the fact that it is outperformed by AR even more remarkable. The comparison is performed for the constant-time replay strategy, which is fair since both AR and VAE-DGR are trained with the same number of samples.
	\par\noindent\textbf{DGR assumptions}
	DGR in general requires that sample be presented to the scholar in the proportions they occur in the data due to the sensitivity of DNNs to class frequencies. Since these frequencies may be unknown in practice, equal class frequencies are usually assumed. DGR also requires the knowledge of all classes that were encountered in past sub-task to compute the correct mixing ratio for the current one.
	\par\noindent\textbf{AR assumptions}
	The fundamental assumption of adiabatic additions is actually not a hard requirement. If additions are large w.r.t. existing knowledge, new data will usually involve more samples, leading to more samples that are replayed by AR.
	Of course it is theoretically possible that a small set of new samples will overlap strongly with all of the previously learned knowledge, but the chances of this happening in practice are virtually zero. \Cref{fig:gen_samples_loglik_plot} b) shows small overlaps for the experiments described here. If we require that all sub-tasks have roughly the same number of samples, then the adiabatic assumption will be fulfilled automatically as more and more sub-tasks are added.
	\par\noindent\textbf{Selective updating in AR}
	Selective updating is an automatic consequence of the update rule for GMMs, which is based on a Mahalanobis-type distance defined by centroids $\vmu_k$ and covariance matrices $\mSigma_k$ of GMM components. As shown in \cite{gepperth2021gradient}, the negative GMM log-likelihood of a single sample, $-\log \sum_k \pi_k \mathcal N(\vx; \vmu_k,\mSigma_k)$ is strongly dominated by a single GMM component $k^*$, giving $-\log (\pi_{k^*} \mathcal N(\vx; \vmu_{k^*}, \mSigma_{k^*}))$. Executing the log yields the expression $(\vx-\vmu_{k^*})^T\mSigma_{k^*}(\vx - \vmu_{k^*})$, showing that the loss is dominated by the best-matching GMM component, which is the component that is adapted as a consequence.
	\par\noindent\textbf{Forgetting} 
	The use of probabilistic models such as GMMs has another interesting consequence, namely the ability to control forgetting.
	Following \cite{zhou2022fortuitous}, forgetting can be a beneficial functionality, which is simple to control in GMMs by eliminating certain components without impacting the remaining ones. 
	
	\par\noindent\textbf{Problems of VAE-DGR}
	In DGR, generator and solver are taken to be DNNs, which are vulnerable to CF. 
	As such, it is imperative to train them on all of the previously acquired knowledge, in correct class proportions.
	This causes the linear growth of replayed samples as observed in \cref{fig:gen_samples_loglik_plot}.
	\par\noindent\textbf{A word on GANs} 
	GANs, although in principle capable of generating high-quality samples, are well-known to require considerable data-dependent tuning to perform well and, e.g., avoid mode collapse (see, .e.g, \cite{thanh2020catastrophic}). We therefore decided not to rely on GANs for DGR. Our choice of using VAEs for DGR has been supported by other studies, e.g., \cite{lesort2019marginal}, using VAEs with comparable CL performance to GANs. Since in CL, all sub-tasks except the first are unknown, we can use data from the first sub-task only to tune GANs structure and hyper-parameters without violating CL conventions. In contrast, VAEs were shown in \cite{dzemidovich2022empirical} and in the present study to be robust to problem like mode collapse, which makes them more suitable for CL.
	\\[3pt]
	Since this is a proof-of-concept study, AR in the presented form has limitations as well:
	\par\noindent\textbf{Scalability/capacity}
	Obviously, the densities that can be learned by a single GMM layer as employed in AR is limited. It seems plausible to assume that datasets like SVHN or CIFAR may be too complex for AR in its current state. As in DNNs, a feasible way to extend representational capacities of GMMs without exploding resource requirements is to construct deep convolutional hierarchies. Such DCGMM models were proposed in \cite{gepperth2021new}, and we intend to use them as drop-in replacements in AR, such that more complex densities can be modeled. 
	\par\noindent\textbf{Initial annealing radius tuning}
	AR contains a few technical details that can be hard to tune, like the initial annealing radius $r_0$ when re-training with new sub-tasks. We used a single value for in all experiments, but performance is sensitive to this choice, since it represents a trade-off between new data acquisition and knowledge retention. Therefore, we intend to develop an automated control strategy for this parameter to facilitate experimentation.
	\section{Conclusion}
	We believe that continual learning (CL) holds the potential to spark a new machine learning revolution, since it allows, if it could be made to work in practice, the training of models over long times. In this study, we show a proof-of-concept for CL that operates at a time complexity that is independent of the amount of previously acquired knowledge, which is something we also observe in humans. Clearly, more complex problems must be investigated under less restricted conditions, and a comprehensive comparison to other CL methods (EWC, LwF, GEM) will be provided in future work. Above all, we believe that more complex probabilistic models such as deep convolutional GMMs must be employed to adequately describe more complex data distributions.	%
	\bibliography{collas2023_conference.bib}
	\bibliographystyle{collas2023_conference}
	\newpage
	\appendix
	\section{Appendix}
	\subsection{Experimental Evaluation: Additional Metrics}
	%
	\begin{table*}[h]
		\tiny
		\centering
\begin{tabular}{ c c | c | c | c | c | c | c | c | c | c | c | c | c | c | c | c | c | c | c }
			& \multicolumn{18}{c}{\large{\textbf{Forgetting Measures}}} \\[12pt] 
			\multirow{2}{*}{\large{CIL-P.}} &                
			& \multicolumn{6}{|c|}{{\large \textbf{AR}}} & \multicolumn{6}{|c|}{{\large \textbf{DGR(c.)}}} & \multicolumn{6}{|c}{{\large \textbf{DGR(b.)}}} \\[6pt]
			& \large{$F_{i,j}$}
			& $T_1$ & $T_2$ & $T_3$ & $T_4$ & $T_5$ & $T_6$ 
			& $T_1$ & $T_2$ & $T_3$ & $T_4$ & $T_5$ & $T_6$
			& $T_1$ & $T_2$ & $T_3$ & $T_4$ & $T_5$ & $T_6$ \\[3pt]
			\hline      
			\multirow{6}{8em}{{\large F-MNIST D5-$1^5$A}}
			& $T_1$
			& \multicolumn{1}{c}{} 
			& \multicolumn{1}{c}{$.00$} 
			& \multicolumn{1}{c}{$.06$}     
			& \multicolumn{1}{c}{$.06$}
			& \multicolumn{1}{c}{$.07$}     
			& \multicolumn{1}{c|}{$.06$} 
			& \multicolumn{1}{c}{} 
			& \multicolumn{1}{c}{$.00$} 
			& \multicolumn{1}{c}{$.06$}     
			& \multicolumn{1}{c}{$.06$}
			& \multicolumn{1}{c}{$.07$}     
			& \multicolumn{1}{c|}{$.06$} 
			& \multicolumn{1}{c}{} 
			& \multicolumn{1}{c}{$.03$} 
			& \multicolumn{1}{c}{$.42$}     
			& \multicolumn{1}{c}{$.09$}
			& \multicolumn{1}{c}{$.12$}     
			& \multicolumn{1}{c}{$.09$}                   
			\\
\end{tabular}
		\caption{Forgetting results for AR and DGR(b./c.) on Fashion-MMNIST and E-MNIST. Severe Forgetting is marked as red for accuracy drops of ($>50\%$). Negative Forgetting/Positive Backwards-Transfer is marked as green. All values are averaged over $N=10$ experiment runs.
			\label{tab:app_forgetting_mats}
		}
	\end{table*}
	%
	\begin{table*}[h]
		\tiny
		\centering
		\begin{tabular}{ c c | c | c | c | c | c | c | c | c | c | c | c | c | c | c | c | c | c | c }
			& \multicolumn{18}{c}{\large{\textbf{Accuracy Measures: Adiabatic Replay (AR)}}} \\[12pt] 
			\multirow{2}{*}{\large{CIL-P.}} &                
			& \multicolumn{6}{|c|}{{\large \textbf{MNIST}}} & \multicolumn{6}{|c|}{{\large \textbf{Fashion-MNIST}}} & \multicolumn{6}{|c}{{\large \textbf{E-MNIST}}} \\[3pt]
			& \large{$a_{i,j}$}
			& $T_1$ & $T_2$ & $T_3$ & $T_4$ & $T_5$ & $T_6$ 
			& $T_1$ & $T_2$ & $T_3$ & $T_4$ & $T_5$ & $T_6$
			& $T_1$ & $T_2$ & $T_3$ & $T_4$ & $T_5$ & $T_6$ \\[3pt]
			\hline    
			\multirow{6}{*}{{\large \textbf{D5-$1^5$A}}}
			& $T_{base}$
			& \multicolumn{1}{c}{$.50$} 
			& \multicolumn{1}{c}{$.55$} 
			& \multicolumn{1}{c}{$.63$}     
			& \multicolumn{1}{c}{$.70$}
			& \multicolumn{1}{c}{$.73$}     
			& \multicolumn{1}{c|}{$.76$} 
			& \multicolumn{1}{c}{$.41$} 
			& \multicolumn{1}{c}{$.50$} 
			& \multicolumn{1}{c}{$.52$}     
			& \multicolumn{1}{c}{$.56$}
			& \multicolumn{1}{c}{$.64$}     
			& \multicolumn{1}{c|}{$.70$} 
			& \multicolumn{1}{c}{} 
			& \multicolumn{1}{c}{} 
			& \multicolumn{1}{c}{}     
			& \multicolumn{1}{c}{}
			& \multicolumn{1}{c}{}     
			& \multicolumn{1}{c}{}                         
			\\[3pt]
			& $T_1$
			& \multicolumn{1}{c}{$.98$} 
			& \multicolumn{1}{c}{$.96$} 
			& \multicolumn{1}{c}{$.96$}     
			& \multicolumn{1}{c}{$.95$}
			& \multicolumn{1}{c}{$.91$}     
			& \multicolumn{1}{c|}{$.85$} 
			& \multicolumn{1}{c}{$.82$} 
			& \multicolumn{1}{c}{$.82$} 
			& \multicolumn{1}{c}{$.76$}     
			& \multicolumn{1}{c}{$.76$}
			& \multicolumn{1}{c}{$.75$}     
			& \multicolumn{1}{c|}{$.75$} 
			& \multicolumn{1}{c}{} 
			& \multicolumn{1}{c}{} 
			& \multicolumn{1}{c}{}     
			& \multicolumn{1}{c}{}
			& \multicolumn{1}{c}{}     
			& \multicolumn{1}{c}{}                           
			\\
\end{tabular}
		\caption{Accuracy matrices for AR. $T_{base}$ was added to showcase performance on the baseline test set, see \cref{sec:exppeval} for detailed information about the evaluation of metrics. All values are averaged over $N=10$ experiment runs.}
		\label{tab:app_ar_accmatrix}
	\end{table*}
	%
	\begin{table*}[h]
		\tiny
		\centering
		\begin{tabular}{ c c | c | c | c | c | c | c | c | c | c | c | c | c | c | c | c | c | c | c }
			& \multicolumn{18}{c}{\large{\textbf{Accuracy Measures: DGR (constant-time setting)}}} \\[12pt] 
			\multirow{2}{*}{\large{CIL-P.}} &                
			& \multicolumn{6}{|c|}{{\large \textbf{MNIST}}} & \multicolumn{6}{|c|}{{\large \textbf{Fashion-MNIST}}} & \multicolumn{6}{|c}{{\large \textbf{E-MNIST}}} \\[3pt]
			& \large{$a_{i,j}$}
			& $T_1$ & $T_2$ & $T_3$ & $T_4$ & $T_5$ & $T_6$ 
			& $T_1$ & $T_2$ & $T_3$ & $T_4$ & $T_5$ & $T_6$
			& $T_1$ & $T_2$ & $T_3$ & $T_4$ & $T_5$ & $T_6$ \\[3pt]
			\hline  
			\multirow{6}{*}{{\large \textbf{D5-$1^5$A}}}
			& $T_{base}$
			& \multicolumn{1}{c}{$.51$} 
			& \multicolumn{1}{c}{$.59$} 
			& \multicolumn{1}{c}{$.67$}     
			& \multicolumn{1}{c}{$.70$}
			& \multicolumn{1}{c}{$.60$}     
			& \multicolumn{1}{c|}{$.63$} 
			& \multicolumn{1}{c}{$.45$} 
			& \multicolumn{1}{c}{$.54$} 
			& \multicolumn{1}{c}{$.41$}     
			& \multicolumn{1}{c}{$.56$}
			& \multicolumn{1}{c}{$.59$}     
			& \multicolumn{1}{c|}{$.63$} 
			& \multicolumn{1}{c}{} 
			& \multicolumn{1}{c}{} 
			& \multicolumn{1}{c}{}     
			& \multicolumn{1}{c}{}
			& \multicolumn{1}{c}{}     
			& \multicolumn{1}{c}{}                         
			\\[3pt]
			& $T_1$
			& \multicolumn{1}{c}{$.99$} 
			& \multicolumn{1}{c}{$.97$} 
			& \multicolumn{1}{c}{$.95$}     
			& \multicolumn{1}{c}{$.83$}
			& \multicolumn{1}{c}{$.50$}     
			& \multicolumn{1}{c|}{$.40$} 
			& \multicolumn{1}{c}{$.91$} 
			& \multicolumn{1}{c}{$.87$} 
			& \multicolumn{1}{c}{$.43$}     
			& \multicolumn{1}{c}{$.63$}
			& \multicolumn{1}{c}{$.53$}     
			& \multicolumn{1}{c|}{$.53$} 
			& \multicolumn{1}{c}{} 
			& \multicolumn{1}{c}{} 
			& \multicolumn{1}{c}{}     
			& \multicolumn{1}{c}{}
			& \multicolumn{1}{c}{}     
			& \multicolumn{1}{c}{}                           
			\\
\end{tabular}
		\caption{Accuracy matrices for DGR (constant-time setting). $T_{base}$ was added to showcase performance on the baseline test set, see \cref{sec:exppeval} for detailed information about the evaluation of metrics. All values are averaged over $N=10$ experiment runs.}
		\label{tab:app_dgr-c_accmatrix}
	\end{table*}
	%
	\begin{table*}[h]
		\tiny
		\centering
		\begin{tabular}{ c c | c | c | c | c | c | c | c | c | c | c | c | c | c | c | c | c | c | c }
			& \multicolumn{18}{c}{\large{\textbf{Accuracy Measures: DGR (balanced setting)}}} \\[12pt] 
			\multirow{2}{*}{\large{CIL-P.}} &                
			& \multicolumn{6}{|c|}{{\large \textbf{MNIST}}} & \multicolumn{6}{|c|}{{\large \textbf{Fashion-MNIST}}} & \multicolumn{6}{|c}{{\large \textbf{E-MNIST}}} \\[3pt]
			& \large{$a_{i,j}$}
			& $T_1$ & $T_2$ & $T_3$ & $T_4$ & $T_5$ & $T_6$ 
			& $T_1$ & $T_2$ & $T_3$ & $T_4$ & $T_5$ & $T_6$
			& $T_1$ & $T_2$ & $T_3$ & $T_4$ & $T_5$ & $T_6$ \\[3pt]
			\hline      
			\multirow{6}{*}{{\large \textbf{D5-$1^5$A}}}
			& $T_{base}$
			& \multicolumn{1}{c}{$.51$} 
			& \multicolumn{1}{c}{$.59$} 
			& \multicolumn{1}{c}{$.68$}     
			& \multicolumn{1}{c}{$.77$}
			& \multicolumn{1}{c}{$.78$}     
			& \multicolumn{1}{c|}{$.83$} 
			& \multicolumn{1}{c}{$.45$} 
			& \multicolumn{1}{c}{$.54$} 
			& \multicolumn{1}{c}{$.44$}     
			& \multicolumn{1}{c}{$.63$}
			& \multicolumn{1}{c}{$.70$}     
			& \multicolumn{1}{c|}{$.77$} 
			& \multicolumn{1}{c}{} 
			& \multicolumn{1}{c}{} 
			& \multicolumn{1}{c}{}     
			& \multicolumn{1}{c}{}
			& \multicolumn{1}{c}{}     
			& \multicolumn{1}{c}{}                         
			\\[3pt]
			& $T_1$
			& \multicolumn{1}{c}{$.99$} 
			& \multicolumn{1}{c}{$.98$} 
			& \multicolumn{1}{c}{$.98$}     
			& \multicolumn{1}{c}{$.97$}
			& \multicolumn{1}{c}{$.84$}     
			& \multicolumn{1}{c|}{$.78$} 
			& \multicolumn{1}{c}{$.91$} 
			& \multicolumn{1}{c}{$.87$} 
			& \multicolumn{1}{c}{$.49$}     
			& \multicolumn{1}{c}{$.81$}
			& \multicolumn{1}{c}{$.78$}     
			& \multicolumn{1}{c|}{$.77$} 
			& \multicolumn{1}{c}{} 
			& \multicolumn{1}{c}{} 
			& \multicolumn{1}{c}{}     
			& \multicolumn{1}{c}{}
			& \multicolumn{1}{c}{}     
			& \multicolumn{1}{c}{}                           
			\\
\end{tabular}
		\caption{Accuracy matrices for DGR (balanced setting). $T_{base}$ was added to showcase performance on the baseline test set, see \cref{sec:exppeval} for detailed information about the evaluation of metrics. All values are averaged over $N=10$ experiment runs.}
		\label{tab:app_dgr-b_accmatrix}
	\end{table*}
\end{document}

%% file: math_commands.tex
\usepackage{amsmath,amsfonts,bm}

\def\eqref#1{equation~\ref{#1}}

\def\1{\bm{1}}

\def\vmu{{\bm{\mu}}}

\def\vo{{\bm{o}}}

\def\vx{{\bm{x}}}

\def\mW{{\bm{W}}}

\def\mSigma{{\bm{\Sigma}}}

\DeclareMathAlphabet{\mathsfit}{\encodingdefault}{\sfdefault}{m}{sl}
\SetMathAlphabet{\mathsfit}{bold}{\encodingdefault}{\sfdefault}{bx}{n}

